\documentclass{article}

\usepackage{arxiv}

\usepackage[utf8]{inputenc} % allow utf-8 input
\usepackage[T1]{fontenc}    % use 8-bit T1 fonts
\usepackage{hyperref}       % hyperlinks
\usepackage{url}            % simple URL typesetting
\usepackage{booktabs}       % professional-quality tables
\usepackage{amsfonts}       % blackboard math symbols
\usepackage{nicefrac}       % compact symbols for 1/2, etc.
\usepackage{microtype}      % microtypography
\usepackage{lipsum}
\usepackage{graphicx}
\usepackage{subcaption}
\usepackage{float}

\usepackage{amsmath,amsfonts,amssymb,amsthm}

\title{Evolutionary self-replication as a mechanism for producing artificial intelligence}

\author{
 Samuel Schmidgall\\
  U.S. Naval Research Laboratory
  \and \textbf{Joseph Hays} \\
 U.S. Naval Research Laboratory \\
}

\begin{document}
\maketitle
\begin{abstract}

%Artificial intelligence is
%Can artificial intelligence emerge without any form of objective optimization based solely on life, death, and self-replication?
%Common robotic learning environments are re-defined in terms of natural selection

%Where were you during the rise of self-replicators?

%Is it possible to design robots and other machines that can reproduce and evolve?

Can reproduction alone in the context of survival produce intelligence in our machines? In this work, self-replication is explored as a mechanism for the emergence of intelligent behavior in modern learning environments. By focusing purely on survival, while undergoing natural selection, evolved organisms are shown to produce meaningful, complex, and intelligent behavior, demonstrating creative solutions to challenging problems without any notion of reward or objectives. Atari and robotic learning environments are re-defined in terms of natural selection, and the behavior which emerged in self-replicating organisms during these experiments is described in detail.

%This work demonstrates that self-replication with random mutation, in the context of survival, is sufficient for producing intelligence.

%re-define common RL benchmarks to fit natural selection

%apply to environments and describe the interesting behavior that emerges

% As a product of their mere existence, a diversity of organisms emerge, each suited for their particular domain. By choosing these domains as modern control problems, a suite of tasks are solved deriving from a common ancestor, and producing a phylogeny of diverse organisms, including the emergence of a species of general problem solvers which optimize for performance across all tasks rather than one in a particular niche.

\end{abstract}

%\textit{“Evolution will occur whenever and wherever three conditions are met: replication, variation (mutation), and differential fitness (competition)" (Dennett)}
%How simple can a cell be and still be considered as living? The answer depends on what we consider to be the essential properties of life. Defining life is notoriously difficult; its very diversity resists the confines of any compact definition. An operational approach focuses on identifying simple cellular systems that are both autonomously replicating and subject to Darwinian evolution (Szostak et al, 2001: 387).}

\section{Introduction}

% In the abstract, evolution is a relatively simple concept whereby organisms better suited for their environment tend to survive longer and produce more offspring, which are similar but not identical to themselves.
%through a spiral of ever-growing complexity
For decades it has been the primary strategy of Artificial Intelligence (AI) research to hand-craft rules, logic, architectures, and objectives for our machines to closely follow, making it difficult for open-endedness to emerge (\cite{NILSSON199131,rewardisenough, hayes1985rule, resnick1997recommender, hearst1998support}). Major progress in AI has followed a pattern of lessening manual design, and allowing the machine to discover its own solutions -- to explore. Perhaps evolution is successful precisely because there is so much room for what is defined as fit through the lens of life and death. Why is it that we are not harnessing this approach to bring life, creativity, and intelligence to our machines?

Exploration seems to be the nature of innovation; in fact, unguided exploration, independent of objectives has shown to be an effective means of producing intelligence itself (\cite{noveltyescape, DBLP:journals/corr/abs-1712-06560, eysenbach2018diversity, Lehman2011}). It is well known that objective-based optimization can be deceptive, ironically preventing the optimal solution according to the same objective being optimized from being reached (\cite{stanley2015greatness, abandonobj, mouret2015illuminating, 10.3389/frobt.2016.00040}). Significant manual effort goes into the design of reward, hyper-parameters, architectures, and learning algorithms, whereas, realistically, such nuances could be discovered by an algorithm itself. The paradigm of AI-generating algorithms (AI-GAs) argues that a more effective method of producing AI is to introduce a process whereby the AI learns how to design and improve itself (\cite{clune2020aigas}). In reality, this is precisely the process with which biological organisms have developed. %, with evolution as the top-level design process. 

It is easy to neglect the idea that evolution emerges as a \textit{property of reproducing organisms} under the pressure of survival; as in, it may look as if there is an overarching algorithm being \textit{applied to} the organisms, whereas, it is rather that the reproductive organisms themselves, under the pressure of natural selection, undergo evolution. While this may seem like just a matter of semantics, conceptualizing evolution in this way alters the way in which algorithms seeking to harness the power of evolution are executed. Evolution is not simply applied to organisms, rather, the organisms themselves are designed in such a way that they evolve naturally. This begs the question, how can evolution be implemented in this way to produce organisms which are interesting and of use for us? This work answers that question by abstracting the properties of biological self-replication and formulating modern learning problems in terms of natural selection: life and death. In this way, interesting behavior on a series of robotic learning problems, as well as several Atari games, is observed to emerge based solely on life, death, and replication in the absence of reward.

\section{Related Work}

The division between evolutionary computation and evolutionary self-replication is important to recognize. Evolutionary computation operates as follows: (1) generate an initial population of random individuals, (2) assign each individual a fitness based on a defined reward or objective function, (3) use the obtained fitness evaluations to perform selection or parameter optimization, (4) create a new population by varying the previous population, (5) repeat from 2. Whereas, evolutionary self-replication typically operate as follows: (1) generate initial organism(s), (2) organisms must meet conditions for survive or cease to exist, and (3) organisms which survive longer tend to replicate more often.

From a biological perspective, evolutionary algorithms have been argued to lack crucial components of the evolutionary process (\cite{eiben2015evolutionary}). One primary difference is that the execution of evolutionary algorithms is centralized, with synchronized manual selection periods in the form of generations, unlike the parallel and decentralized process of natural evolution. There is also no notion of spatial relationships between individuals, and population sizes are kept constant through highly synchronous timing and organization, whereas natural evolution is spatially organized, and population sizes vary according to birth and death events, occasionally going extinct. Additionally, evolutionary algorithms have no self-referential notion of life, death, or reproduction -- such processes are \textit{applied} to individuals. Evolutionary self-replicators more closely resemble the process of natural evolution. The fundamental difference between evolutionary algorithms and evolutionary self-replicators, is that evolutionary algorithm methods undergo a structured form of \textit{artificial selection}, whereas evolutionary self-replicators undergo \textit{natural selection} (\cite{ray1992evolution, channon1997artificial, channon1998perpetuating}).

Evolutionary self-replicating programs have a rich history in the field of Artificial Life (ALife) and Artificial Intelligence (AI). The first organisms to undergo digital evolution were from the program Darwin, created in Bell Labs in 1961. In the game Darwin, computer programs compete with each other by trying to stop the execution of other programs. Further advancements includes work on Tierra (\cite{ray1992evolution, ray2001measures, ray1991approach}) where it was observed for the first time that evolutionary self-replicating programs evolve meaningfully, and toward higher complexity. Work in Tierra found that the emergence of tens of thousands of unique programs is possible from a single ancestral organism. Avida extends work in Tierra under a set of different dynamics (\cite{ofria2004avida, lenski2003evolutionary, adami2000evolution, lenski1999genome}). Avida experiments demonstrated a capability of digital organisms to evolve necessary structure to solve logic puzzles in order to gain energy in the form of computation time (\cite{lenski1999genome}). These discoveries show that evolutionary self-replicators are creative, surprising, and most of all, intelligent. % (\cite{lehman2020surprising}) 

While these worlds are rich in potential and discovery, it is often the case that we'd like our organisms to solve more concrete, applicable, or interesting problems. Toward this end, we present three scenarios of environment reformulation based on survival, foraging, and survival-foraging, which address these expectations.

%When an organism demands existence above all else, finding loop-holes for continued life are to be expected. The intelligence of the environment should be such that existence is both desired, as well as routinely performing some productive task. Toward this end, we present three scenarios of environment reformulation based on survival, foraging, and survival-foraging, which address these expectations, and a replication process resembling bacterial evolution is explored to better handle stochastic environments.

%\url{https://github.com/SamuelSchmidgall/EvolutionarySelfReplication}

\section{Evolutionary Self-Replication}

\subsection{A simple evolutionary self-replicator}

A simple evolutionary self-replicator is introduced whose survival is defined by maintaining an \textit{equilibrium} and whose self-reproductive success is defined probabilistically over the course of the replicator's lifetime\footnote{To allow for more descriptive commentary of the results in this section the implementation and code is elsewhere discussed in significant depth at \\ \url{https://github.com/SamuelSchmidgall/EvolutionarySelfReplication} and in the supplementary material.}. Initially, a finite grid of empty 1-dimensional cells are instantiated (Figure \ref{fig:sequence}). From here, a "basic organism" emerges, from which the specific qualities of this basic organism depends on the environment and definition of the self-replicator; in the simplest case, a small neural network which initially produces random behavior emerges. To retain the utmost simplicity, the actual process of replication is abstracted\footnote{As in, for simplicity, the process of replication is not evolved with the organism but is a built in mechanism.}, and mutations are introduced through random noise. The environment that the organism finds itself in defines survival, life, and death, and to retain existence, the organism must be adapted for survival. If the organism encounters death, then it ceases to exist and is removed from the 1-dimensional grid.

Replication on the 1-dimensional grid works as follows: a) at every environment timestep each organism has a probability of replication, b) if an organism successfully replicates, a mutated copy of itself is randomly produced in either one of the two neighboring cells, c) if the selected cell is occupied then replication does not occur. In this way, once the cells become increasingly occupied, replication on the 1-dimensional grid becomes a competition for longevity. There is no reward: either the organism lives or it dies. The longer an organism is able to survive, the more likely it is to replicate; the copies replicated from organisms with longer lifespans are expected to be more likely to survive than those derived from organisms with shorter lifespans, containing similar, but not identical, properties to their ancestral organisms. Evolution occurs without guidance and without explicit selection, instead, natural selection.

\begin{figure}
    \centering
    \includegraphics[width=.48\linewidth]{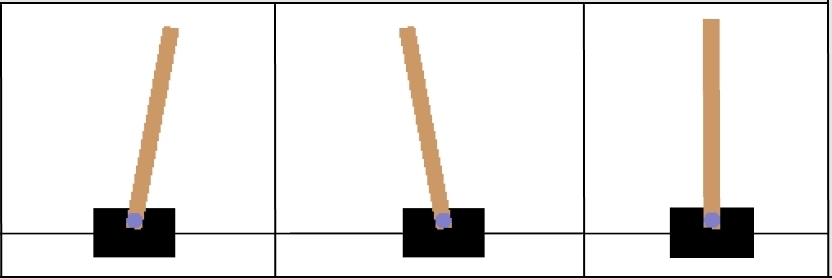}
    \caption{Three configurations of the CartPole-survival environment. (Left) The pole is falling to the right, (Center) the pole is falling to the left, (Right) the pole is balanced. Once the pole falls beyond a certain limit the organism dies.}
    \label{fig:cartpole}
\end{figure}

\textbf{CartPole-survival:} The first environment in consideration, which has a natural definition of life and death, is CartPole-survival (Figure \ref{fig:cartpole}). In this environment, a pole is placed in the center of a cart, from which the organism may apply force to the cart on either the right or left side. If the top of the pole falls beyond a certain threshold off of the cart, the organism dies. Until a death state is reached, the organism continues existence.

\begin{figure}
    \centering
    \includegraphics[width=.85\linewidth]{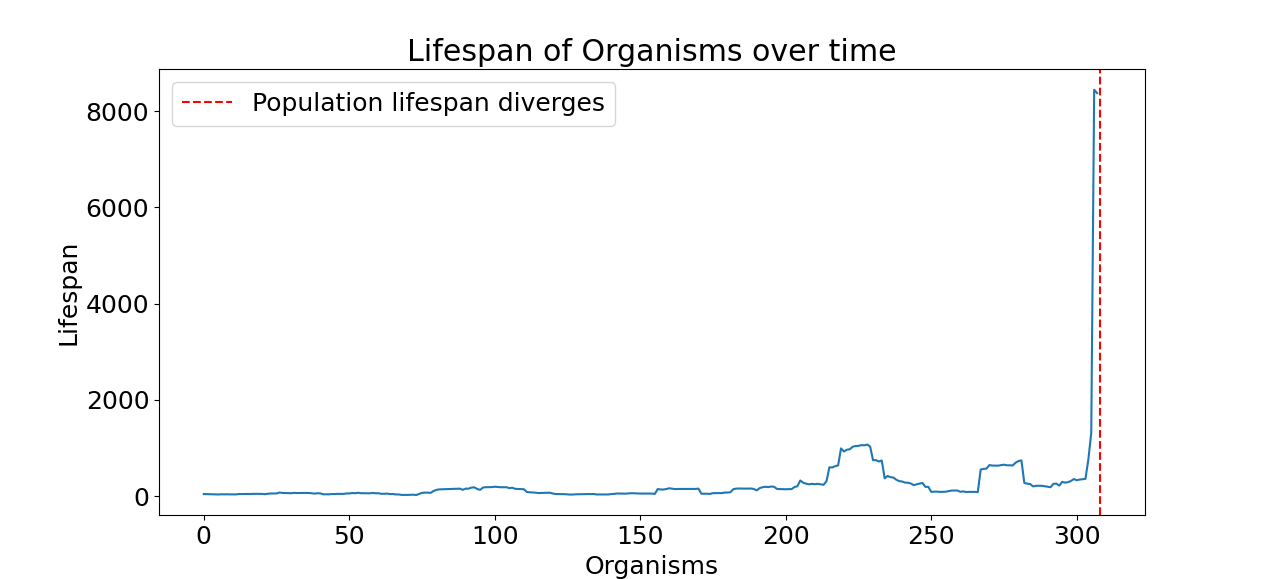}
    \caption{Average lifespan of evolutionary self-replicating organisms over time in the CartPole-survival environment. After around 300 organisms the population lifespan diverges and a "perfect" solution is reached where the organism never reaches the death-state. This behavior is replicated to the other cells and an immortal population is produced.}
    \label{fig:population_div}
\end{figure}

CartPole-survival serves a good initial environment since it requires few modifications to introduce natural selection. In this environment, life is defined as retaining balance within a certain interval of angles, and death occurs if the pole goes outside of that range. Each organism's behavior is produced by a fully-connected recurrent neural network with 32 hidden neurons, from which random mutations perturb synaptic weights. Running this experiment in a size 32 1-dimensional grid produces the lifespan dynamics in Figure \ref{fig:population_div}. Initially, the first 200 organisms survive an average of around 100 timesteps, until the population lifespan average suddenly rises to 1500 timesteps near the 225th organism. This value fluctuates until after the 300th organism, where an infinite lifespan organism emerges. This "immortal" organism's behavior is quickly reproduced into neighboring cells, and soon every organism in the population has a perfect solution and organism death ceases to occur.

\subsection{Re-defining robotic learning}

To solve the familiar suite of modern robotic-learning tasks, first, the tasks must be re-defined in a way which is suitable for the emergence of natural selection: life and death. Three approaches toward this re-definition are considered in this section: survival, foraging, and a survival-foraging.

The environments presented were selected prior to experimentation, and the solutions were not cherry-picked; the first obtained solutions are presented for all tasks. In many environments, the organisms produced through evolutionary self-replication display the utmost creativity in their solutions, often exploiting nuances in the environment itself to increase odds of survival, which are behaviors inherited by successors. Small neural networks are used so that the organisms are pressured not to simply memorize information, as many over-parameterized models tend to do. Each task uses a single hidden-layer fully-connected recurrent neural network with 32 neurons in the hidden-layer.

While robotic learning may not be the most intuitive application of evolutionary self-reproduction, as biological organisms started from the ground-up, inventing their own problems, and robotic learning applications are born tabla-rasa with a fixed morphology, it is exactly for that reason why robotic learning serves as a good example. The origins of life \textit{need not} be re-created to harness the incredible capabilities of evolution.

\textbf{Survival}\\
Survival-based environments, as was seen with the CartPole-survival environment, require the maintenance of some sort of state which is defined by the environment. Once a certain condition is met, a death-state, then the organism is terminated and ceases to exist, preventing further reproduction. In survival-based environments, it is the sole avoidance of external death-states that enable an organisms to survive, independent of additional criterion. As is seen in the following survival experiments, this sole focus on survival may, though not always, lead to organisms performing clever environment exploitation, where death-states are avoided by manipulating flaws in the environment's design.

\begin{figure}
    \centering
    \includegraphics[width=.85\linewidth]{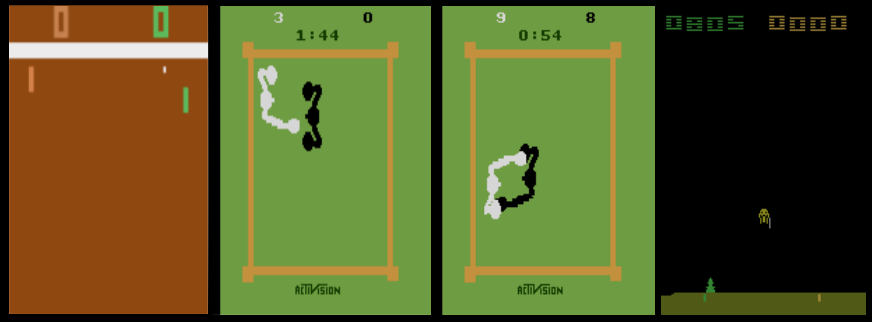}
    \caption{Four survival environments, Pong-survival (left), Boxing-survival-avoidance (middle left), Boxing-survival-fight (middle right), and Space-Invaders-survival (right). Pong paddle is depicted in its "exploited" position, with the opposing paddle (orange) constantly returning the ball to the exact same position. Boxing-survival-avoidance corner position, with the arm constantly punching for opponent deterrence. The more aggressive Boxing-survival-fight organism running down the clock without scoring too many points. Space-Invaders-survival, hiding from the last remaining alien to stay alive longer.}
    \label{fig:survival}
\end{figure}

For the Atari environments presented, Random Access Memory (RAM) states were selected for use as sensory input because the natural environment of digital organisms is digital data, and hence it is useful to show that RAM states may be used meaningfully. RAM state training is much less studied compared to image-state information, perhaps as a result for being more difficult to learn because of their discrete nature \cite{atari_ram}. Nonetheless, the solutions to the Atari environments presented here utilize RAM.

\textbf{\textit{Boxing-survival:}} The Boxing-survival environment has been adapted from an existing Atari game \cite{ALE}, in this case, Boxing. In this environment, two players, our organism and a pre-defined AI, are placed in a boxing ring where actions include moving around and punching with either fist (Figure \ref{fig:survival}). First individual boxer to 100 points wins, where points are defined by how many punches are landed on the opposing player. The fight has a 2-minute timer, and the original environment ends either when an individual reaches 100 points or when the timer runs out. Converting this environment opened several possibilities in defining death. Two possible definitions of varying difficulty are considered in this experiment; the first environment, Boxing-survival-fight, defines the death-state as the moment the opponent obtains more points, and, among the more difficult possibilities, the second environment, Boxing-survival-avoidance, defines getting punched a \textit{single time} as the death-state. 

In Boxing-survival-fight, the strategy for continued existence is non-trivial. Consider the case of if the organism were to significantly out-box the opposing side. In this scenario, the organism would reach 100 points before the timer runs out, and hence, organisms which did not score quite as quickly, but still managed to do better than the opponent would be more fit for survival. Early successful replicators, which were the most efficient at defeating the opponent quickly, served as a critical catalyst for obtaining a larger population lifespan, but eventually dwindled in favor of more reserved organisms, which did not rush to 100 points; rather, these organisms only obtained points when necessary in favor of dodging punches and waiting for the clock to run out. This ultimately led to a population of organisms which survived the maximum amount of time possible -- the entire 2-minute timer.

The Boxing-survival-avoidance environment is not as forgiving as the survival-fight environment was. This environment does not allow the organism to receive a single hit from the opponent, which creates only a single, and quite challenging, possibility for continued existence: to avoid the enemy's punches completely. The organism discovers that if it stands in one of the boxing-ring corners and rapidly punches its arm opposite the corner, then its opponent will walk right into the punches and automatically back up, move around, and come back, without ever throwing a punch -- an exploitation of the enemy AI design. This way, using a simple yet clever strategy, the organism successfully avoids being hit by its opponent until the timer runs out. Upon initially discovering this strategy, right-arm punches in the corner were chaotic and relentless, without much consideration for timing; however, over time the punches became more well-timed and deliberate as survival time increased.

\textbf{\textit{Pong-survival:}} Like Boxing-survival, the Pong-survival environment adapts the existing Atari Pong game, where two paddles are placed on opposing sides and must deflect a horizontally-travelling ball, obtaining points when the ball passes their adversary's paddle (Figure \ref{fig:survival}). However, since survival environments only deal with avoiding death-states, not point-scoring, this environment is modified where, if a point is scored past the organism's paddle, then it dies. In the Pong-survival environment, by nature of the environment definition, it is in the organism's best interest not to win \textit{or} lose the game; rather, the organism would be most successful to extend the game as long as possible to increase individual odds of reproduction. The strategy developed by the organism population exploits a weakness in the environment's design to attain an infinite lifespan. Initially, the strategy consists of deflecting the ball in a relatively predictable manner, occasionally winning points, but not putting too much stress on the opposing paddle. However, eventually, the organism discovers that, since the opposing paddle is programmed deterministically, there exists a position where the organism paddle may be placed without moving and the opposing paddle will always return to that position. By discovering this paddle position, and learning how to exploit it, the behavior of the self-replicator ultimately leads to an infinite game where no points are scored, and hence, an infinite lifespan.

\textbf{\textit{Space-Invaders-survival:}} Space-Invaders-survival presents a significantly higher degree of complexity compared to the Pong- and Boxing-survival environments. In this environment, the organism is represented by a spaceship at the bottom of the screen moving along the x-axis (Figure \ref{fig:survival}). Aliens stack on top of each other in a row, firing down at the organism, and the organism may destroy enemies by returning fire. There are three destructable defences which the organism can temporarily cover under until the missile attack ultimately destroys them. The goal in the original environment, based on the defined objective function, is to destroy all of the aliens before losing the three allotted lives. This environment naturally translates to survival, since there is already a definition of life and death built into the game.

Like the other environments, the solution to this problem is not trivial, and is quite surprisingly creative. One would imagine the clear strategy is to minimize enemy bombardment by eliminating opposing aliens. However, once the final alien is destroyed, then the game ends, and so does the organism's life. Much like in the Pong-survival environment, it is not in the organism's best interest to actually win the game, rather finding an exploit for an extended lifespan is more desirable. This remains consistent with the organism solution. To obtain a lifespan beyond completing the game, the organism obtains a solution as so to destroy every alien expect for one and keeps the alien alive while skillfully avoiding its attacks. 

\textbf{Foraging}

Not all desirable human-constructed environments lend themselves toward being defined immediately in terms of life and death. Additionally, many practical considerations may require routinely performing some task. Foraging-based environments introduce the notion of a metabolism, whereby the organism must meet some requirement at regular intervals in order to survive. This requirement may be defined by a wide-variety of possibilities, for example: obtaining resources, scoring points in a game, or progressing toward some location.

Foraging tasks tend to be less prone to environment exploitation, as routinely fulfilling some energy-requirement becomes necessary. The introduction of foraging enables behavior which is desirable in the real world, considering physical robots will likely require obtaining energy, resources, or maintenance.

\begin{figure}
    \centering
    \includegraphics[width=.85\linewidth]{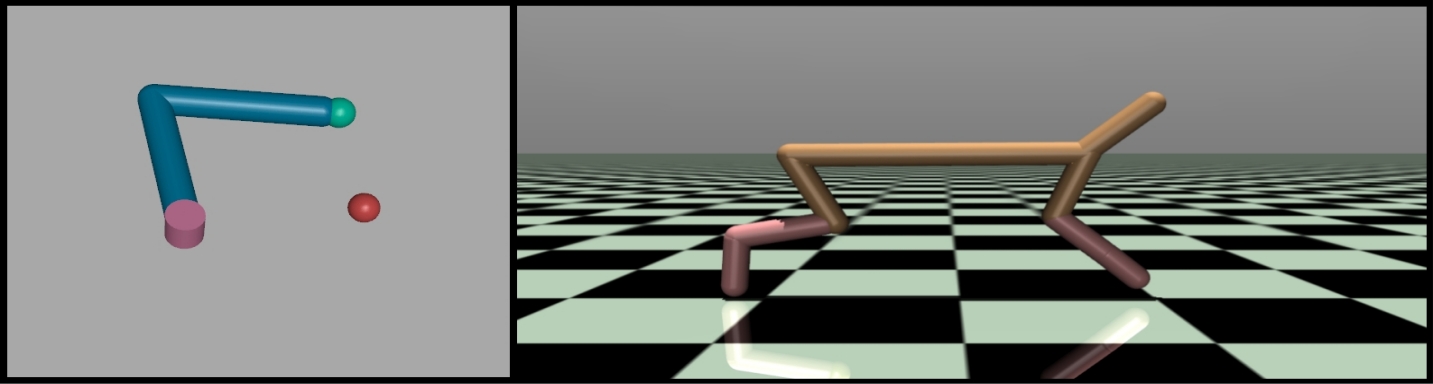}
    \caption{Depiction of the two forager tasks, Reacher-forager (left) and HalfCheetah-forager (right). In the Reacher-forager environment, the goal position is a red sphere, and the tip of the second link is a green sphere. The HalfCheetah-forager environment must make forward progress (right) to obtain energy.}
    \label{fig:foragers}
\end{figure}

\textbf{\textit{Reacher-forager:}} The Reacher-forager environment adapts the original Reacher reinforcement learning benchmark, where a two-link arm resting on a plane is controlled with the task of reaching the tip of the second link onto a goal position. This task may naturally be extended to the foraging domain, by simply requiring that the goal position be reached to obtain energy. Once energy runs out, the organism dies. When a goal position is reached, energy is deposited into the organisms energy storage, which slowly depletes over time, and the goal position is reset.

The solution to this task is quite clever, and goes beyond what would be expected of the organism. Instead of carefully placing the second link in the correct location, the arm develops a more prosperous strategy of rapidly spinning the arm in a circle while expertly compressing and expanding the second link such that all possible goal positions are reached before energy is depleted. Using this solution, the Reacher-forager organism population becomes capable of indefinite existence. By adding an energy penalty to the Reacher's movement, this behavior returns to the expected strategy of precise and slow movement toward goal positions.

\textbf{\textit{HalfCheetah-forager}:} The HalfCheetah-forager environment is adapted from the a common robotic learning benchmark, where a 2-dimensional robotic body, meaning movement is restricted to a plane, shaped like a cheetah is tasked with optimizing for the highest possible velocity. To transform this task into a foraging environment, energy is introduced in the same way as Reacher-forager; however, instead of reaching for target positions, the robotic body obtains energy by making forward movement.%\footnote{Can be thought of as grazing, if you're interested in making a biological analogy.}.

This task leaves less room for creativity, as it is quite straightforward to obtain longevity. The organisms behavior quickly reached a point where the locomotion being produced, and hence the energy being obtained, was sufficient for sustaining longevity. Once a sufficient locomotion pattern was produced, the population's lifespan diverged and the organisms no longer experienced death. Interestingly, the produced locomotion is competitive with the original HalfCheetah objective function when compared to a reinforcement learning based approach.

\textbf{Survival-foraging}

Often times it is useful to incorporate both death-states and metabolism into an environment definition. Both real-world and simulated applications of learning often seek to avoid undesirable states, while simultaneously fulfilling periodic requirements. In this section, several examples of environments which benefit from both survival and forager environment traits are presented, as well as their creative solutions.

\begin{figure}[H]
    \centering
    \includegraphics[width=.75\linewidth]{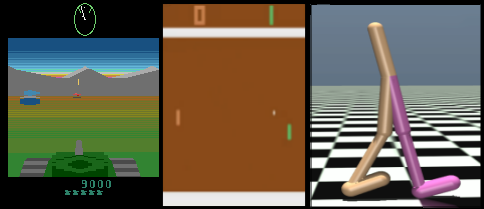}
    \caption{BattleZone-survival-forager (left), Pong-survival-forager (middle), Walker-survival-forager (right)}
    \label{fig:homeoforagers}
\end{figure}

\textbf{\textit{Pong-survival-forager:}} Pong-survival-forager environment extends the original Pong-survival environment by now requiring the organism to score points at regular intervals to stay alive. This prevents the environment exploitation seen in the survival scenario, while simultaneously producing results that are interesting to the researcher; that is, scoring points against an adversary.

The solution found in the Pong-survival-forager environment is very clever, much like the solution found in the Pong-survival environment. Instead of finding a paddle position which leads to an infinite game without any scoring, since this organism requires winning points to stay alive, the evolved organisms finds a paddle position where, once again, exploiting the opposing paddle's behavior, it becomes impossible to lose points while occasionally scoring points on the opposite side. This position exploits the fact that, at the right angle, the ball can bounce off of the \textit{side} of the paddle rather than having to hit the front (Figure \ref{fig:homeoforagers}). This way, the solution obtains the maximum number of points without losing a single point over the course of its lifespan. Despite having no concept of reward, this solution would be the globally optimal solution in an objective-based optimization, hence, state of the art.

\textbf{\textit{Walker-survival-forager}}

The robotic learning benchmark, Walker2d, aims to produce locomotion in a bipedal walker along a restricted plane, much like the HalfCheetah task. The Walker-survival-forager interpretation of this environment defines the upper-body of the biped falling beyond a certain height as the death state, and deposits energy in the organism based on forward progress along the plane. 

To examine the efficacy of evolutionary self-replicators in stochastic settings, this environment has the action output being treated as a probability distribution rather than as deterministic behavior. Since the action domain in this environment is a real vector, the network output is used to represented the mean of a Gaussian distribution, which is sampled to obtain the action applied to the robotic body. To prevent the organism from driving randomness to zero, the standard deviation of this distribution remains fixed.

In the initial set of experiments, in the Walker environment, the evolutionary self-replicators struggled to retain genetic stability, and whenever the overall population lifespan would begin to increase, it would quickly be followed by a sudden crash, occasionally causing an extinction event. A likely candidate cause for the observed instability is that in probabilistic settings, the advantage of a particular set of behaviors may only be adequately evaluated when observed across multiple trials. In policy gradient, genetic, and evolutionary algorithms, this is often addressed by averaging the performance of a single set of parameters over multiple trials before performing updates. In both reality and in this work however, organisms only have one chance at life. 

Biological self-replicators, particularly bacteria, address this issue by having the majority of self-replicators producing identical copies of themselves, whereas only a minority of individuals are mutated \cite{watford2020bacterial}. This concept of probabilistic mutation is introduced to produce much stabler evolutionary dynamics in this task. While bacterial colonies have an incredibly low probability of mutations per replication, for this task, it was found to be sufficient to use a mutation probability of $10\%$. It was also found beneficial for stability to allow replication into any cell on the 1-dimensional grid, rather than just in adjacent ones.

Using these modifications, a high-quality solution to the Walker-survival-forager task is solved by the stochastic self-replicators. Much like in the HalfCheetah environment, the solution is rather trivial, as the task does not allow much room for alternative solutions. However, the population lifespan dynamics were interesting in that instead of the increase in lifespan over time graph resembling a curve, it more so resembled frequent plateaus with occasional jumps in performance. These plateaus were likely responsible for maintaining the parametrically similar population, whereas, the occasional dips from deleterious mutations are quickly out-reproduced, and the beneficial mutations become introduced as the new population average, represented as plateau jumps.

\textbf{\textit{BattleZone-survival-forager:}} BattleZone is an Atari game where the organism, a tank, is placed in an environment filled with other enemy tanks. The tank begins with 5-lives and the game score is based on the number of enemy vehicles destroyed before all lives are lost. In converting this to a survival-forager environment, the survival component includes death after all 5-lives are lost, and the forager component replenishes hunger after destroying an enemy vehicle. When foraging is removed, the organism strategy defaults to avoiding all tanks and driving around aimlessly, since the more vehicles that get destroyed, the higher the game difficulty gets.

While it was indeed not the intention of these organisms to perform competitively with state of the art optimization, and while these organisms were not optimized for the defined objective function, the reward obtained by the BattleZone-survival-forager organism was comparable to many of the highest recorded to the best of our knowledge \cite{horgan2018distributed, fortunato2017noisy, gruslys2017reactor, bellemare2013arcade, yang2019fully}, though several still outperform this \cite{badia2020agent57, kapturowski2018recurrent}. With the destruction of each enemy vehicle producing a reward of 1000, the resulting organism obtained a  median reward of 102000 translating to 102 vehicles destroyed over all five lives. The strategy for obtaining such performance is relatively simple to observe, as the tank skillfully dodges enemy fire while returning fire and averaging 20 enemy eliminations per life lost. The network structure used to obtain this performance was a small fixed-weight structure with no additional modifications. It is likely that significant improvements in performance could be realized through additional capabilities.

Perhaps this is a glimpse into the potential of natural selection and random mutation without objective guidance as a mechanism for producing not only intelligence itself, but also intelligence which is not restricted to the bounds of human design.

\textbf{How long do these experiments take to get interesting results?}

The duration of each experiment depended largely on the speed of the simulator and the complexity of the task. For example, in the fastest experiment, to obtain lifespan divergence on the simple evolutionary replicator in the CartPole-survival domain it only required around one minute of computation. On the other hand, in the slowest experiment, the results obtained in the BattleZone-survival-forager environment required around one day of computation. Considering these experiments were run on a standard single core processor without any speed acceleration, and considering the typical compute requirements of gradient-based experiments and evolutionary computation experiments, evolutionary self-replicators tend to obtain interesting behavior relatively fast. This was a surprising property, as meaningful evolution in natural systems tends to occur at a very slow pace.

\textbf{What are the limitations?}

Some environments were difficult to solve, or did not translate well toward being defined in terms of natural selection. Perhaps this is a function of time, a product of using a small neural network, a small 1-dimensional grid, or the parameters did not work well. It was noticed that in some environments, small changes in parameters produced drastic changes in behavior even with small mutation sizes, and hence random noise mutations across generations struggled to retain desirable traits. Perhaps this was a consequence of the discrete Atari RAM states, or the way in which action-selection was treated. It was also observed that environments with a high degree of randomness together with short average lifespans made it challenging for advantageous traits to propagate across generations. An additional discussion of the limitations of this approach with in-depth examples is also provided at \url{https://github.com/SamuelSchmidgall/EvolutionarySelfReplication}.

\textbf{A discussion of hyper-parameters}

Hyper-parameters remained relatively consistent across each of the experiments, and did not require much manual tuning. Perhaps the most important aspect of design is the way in which mutations occur, as environments respond differently to changes in neural structure. This typically corresponds to a higher or lower weight mutation magnitude, as was done in this work, but future work may also include localizing weight changes or evolving neural topology. The probability of individual reproduction is loosely defined proportional to the average lifespan of the initial behaviorally random organisms in their given environment. Setting the probability too low will prevent replication from occurring, and too high, while still producing desirable behavior, tends to cause evolution to slow down. Nonetheless, setting the reproductive probability to one, as in reproducing at every timestep, still allows for intelligent and desirable behavior to emerge. In the Walker-survival-forager environment, it was discovered that introducing a probability of mutation, rather than mutating each off-spring, drastically improved long-term stability in probabilistic settings. This mutational probability was not explored in-depth, but was found to work well at lower values such as the 10\% used in the Walker-survival-forager environment.

%\subsection{A comparison of lifespan and survival}

%When high-performing behavioral solutions are capable of being learned through evolutionary strategies or reinforcement learning methods, often it is the case that, since the lifespan of an organism is manually selected before learning takes place, often the organism is not capable of maintaining consistent behavior beyond its defined lifespan. The truncation of lifespan is necessary with these learning methods because multiple lifespans of a single or multiple organisms are compared, and the longer the lifespan is, the more expensive it is to learn. It is common for dynamics in robotic learning benchmarks to span only 1-10 seconds of simulated time\footnote{As in, the dynamics are integrated for 1-10 seconds as would be interpreted in real-time, which may be hundreds to thousands of time-steps depending on the integration time in simulation.}, since more realistic time periods would be computationally intractable. This issue becomes even more drastic when dynamic parameters are introduced in the model. This problem is referred to here as the problem of finite lifespan.

\section{Discussion}

Evolutionary self-replication presents itself as a powerful process toward developing both intelligence and life. It is shown in this work that creative, meaningful, and intelligent behavior may emerge through the simple process of life, death, and replication. It was shown that this capability can be harnessed in modern robotic learning environments to obtain both practical and interesting behavior by re-framing problems in terms of survival by natural selection. The organisms which emerge in these environments occasionally produced solutions of immortality, which enables evolved organisms to naturally operate over indefinite scales. %, which compared with the expected lifespan of solutions found through evolutionary computation and reinforcement learning, do not struggle with long-term instability. The solutions developed by these alternative optimization schemes were shown to struggle with generalizing beyond their trained time horizon, perhaps indicating an inherent difficult in producing longevity.

The simplicity of the approach presented in this work is intentional, so as to lay a foundation and allow significant room for future research. In the work on Tierra, the self-replication process is the central focus, and Avida, both replication and logic functions, whereas in this work, the process of survival in complex learning environments is the central focus, with replication being abstracted to a probabilistic process. An interesting possibility of future work could incorporate the replication dynamics of Avida \cite{ofria2004avida, lenski2003evolutionary, adami2000evolution, lenski1999genome} or Tierra \cite{ray2001measures, ray1991approach} together with the survival dynamics and environments presented in this work. Perhaps the concept of reproduction affects the way intelligence is produced in evolutionary self-replication. Another potential work could synthesize evolutionary self-replication and morphological evolution, which may allow for a more obvious depiction of diversity and phylogeny \cite{kriegman2018morphological, cheney2018scalable}. An interesting property of evolutionary self-replicating systems is that development may affect the survival and reproductive capabilities of the organisms, and perhaps the interplay between development and natural selection may be explored \cite{hahn2012development}. Development could be represented as intra-lifetime learning, and evolutionary self-replication as the meta-level optimization \cite{astor2000developmental, hampton2004evolution}. Finally, the emergence of intelligence in evolutionary self-replication has important implications for physical self-replicators, which have the potential to pioneer universe exploration in the form of self-replicating space probes \cite{borgue2021near, freitas1980self}.

An interesting question to ask is whether evolutionary self-replication is a learning algorithm in the traditional sense. In some ways, it resembles other learning algorithms, and in others, it differs drastically. It seems as if the emergence of intelligence in the case of evolutionary self-replication is a matter of phenomenon rather than that of design; a phenomenon of great intrigue to the development of artificial intelligence and life.

\nocite{polyworld, bentley1999evolution, darwin1909origin, taylor2020rise, garwood2019re, pargellis2001digital, ecologicalRL}

%That the organisms to be optimized act as the artist, their performance, a canvas, and that the researcher, that which grant
%\subsection{Artificial and natural selection}

%\subsection{Evolved evolvability}

%There is no special treatment toward high-performing individuals -- if their off-spring do not inherit their high-performance, then those traits are not selected for, and the lineage dies off. Evolvability is much more desirable than individual performance.

\bibliographystyle{unsrt}
\bibliography{references}

\end{document}